\pdfoutput=1

\documentclass[11pt]{article}

\usepackage[final]{acl}

\usepackage{times}
\usepackage{latexsym}
\usepackage{graphicx}
\usepackage{enumitem}
\usepackage{xcolor,colortbl}
\usepackage{multirow}
\usepackage{multicol}
\usepackage{graphicx}
\usepackage{ulem}
\usepackage{caption}
\usepackage{subcaption}
\usepackage{amssymb}
\usepackage{booktabs}
\usepackage{multibib}
\usepackage{makecell}
\usepackage{wrapfig}
\usepackage{color}
\usepackage{amsmath}
\usepackage{stfloats}
\usepackage[listings]{tcolorbox}
\tcbuselibrary{listings,theorems}
\usepackage[framemethod=TikZ]{mdframed}
\usepackage{etoolbox}
\usepackage{tikz}
\usetikzlibrary{tikzmark,calc}
\usetikzlibrary{arrows.meta}
\usepackage{dashbox}
\usepackage{titlesec}
\usepackage{lipsum}
\usepackage{caption}

\newtcbtheorem[]{prompt}{Prompt}%
{colback=green!5,colframe=black,fonttitle=\bfseries, left=.02in, right=.02in,bottom=.02in, top=.02in}{prompt}

\definecolor{UserExampleBg}{HTML}{FAFDF9}

\newmdenv[
    roundcorner=5pt,
    backgroundcolor=UserExampleBg,
    linecolor=black,
    outerlinewidth=1pt,
    frametitlebackgroundcolor=black,
    frametitlefont={\bfseries\color{white}},
]{user_example}

\AtBeginEnvironment{user_example}{\setlength{\parindent}{0pt}}

\usepackage[T1]{fontenc}

\usepackage[utf8]{inputenc}

\usepackage{microtype}

\usepackage{inconsolata}

\title{Common 7B Language Models Already Possess Strong Math Capabilities}

\author{Chen Li$^{1,4}$, Weiqi Wang$^{2,4}$, Jingcheng Hu$^{3,4}$, Yixuan Wei$^{3,4}$, \\
\textbf{Nanning Zheng}$^{1}$\textbf{,} \textbf{Han Hu}$^{4}$\textbf{,} \textbf{Zheng Zhang}$^{4}$\thanks{Project leader. Chen, Weiqi, Jingcheng and Yixuan are interns at MSRA. GitHub: \href{https://github.com/Xwin-LM/Xwin-LM}{Xwin-Math} This repository will be continually updated.}\textbf{,} \textbf{Houwen Peng}$^{4}$\footnotemark[1] \vspace{0.7em} \\
{$^1$IAIR, Xi'an Jiaotong University} \quad
{$^2$University of Science and Technology of China} \\
{$^3$Tsinghua University} \quad
{$^4$Microsoft Research Asia}
 \\
\small{\texttt{edward82@stu.xjtu.edu.cn}} \quad
\small{\texttt{\{v-weiqiwang, t-jingchu, t-yixuanwei, zhez, houwen.peng\}@microsoft.com}}
\\
\small{\texttt{nnzheng@xjtu.edu.cn}} \quad
\small{\texttt{ancientmooner@gmail.com}} 
}
\begin{document}

\maketitle
\begin{abstract}
Mathematical capabilities were previously believed to emerge in common language models only at a very large scale or require extensive math-related pre-training. This paper shows that the LLaMA-2 7B model with common pre-training already exhibits strong mathematical abilities, as evidenced by its impressive accuracy of 97.7\% and 72.0\% on the GSM8K and MATH benchmarks, respectively, when selecting the best response from 256 random generations. The primary issue with the current base model is the difficulty in consistently eliciting its inherent mathematical capabilities. Notably, the accuracy for the first answer drops to 49.5\% and 7.9\% on the GSM8K and MATH benchmarks, respectively. We find that simply scaling up the SFT data can significantly enhance the reliability of generating correct answers. However, the potential for extensive scaling is constrained by the scarcity of publicly available math questions. To overcome this limitation, we employ synthetic data, which proves to be nearly as effective as real data and shows no clear saturation when scaled up to approximately one million samples. This straightforward approach achieves an accuracy of 82.6\% on GSM8K and 40.6\% on MATH using LLaMA-2 7B models, surpassing previous models by 14.2\% and 20.8\%, respectively. We also provide insights into scaling behaviors across different reasoning complexities and error types.
\end{abstract}

\section{Introduction}
Mathematical capabilities have long been considered so challenging that they are thought to emerge in common language models only at a very large scale. For instance, studies by \cite{wei2022emergent,wei2022chain} suggest that only models with size exceeding 50 billion parameters can attain meaningful accuracy or benefit from chain-of-thought processing on math problems. A strategy to equip smaller language models with mathematical abilities involves creating math-specific base models trained on hundreds of billions of math-related pre-training data \cite{lewkowycz2022solving,azerbayev2023llemma}. However, the accuracy of such models remains modest; for example, Llemma-7B~\cite{azerbayev2023llemma} only achieves 36.4\% on the GSM8K dataset \cite{cobbe2021training} and 18.0\% on the MATH dataset \cite{hendrycks2021measuring}.

\begin{figure}[t]
    \centering
    \includegraphics[width=0.48\textwidth]{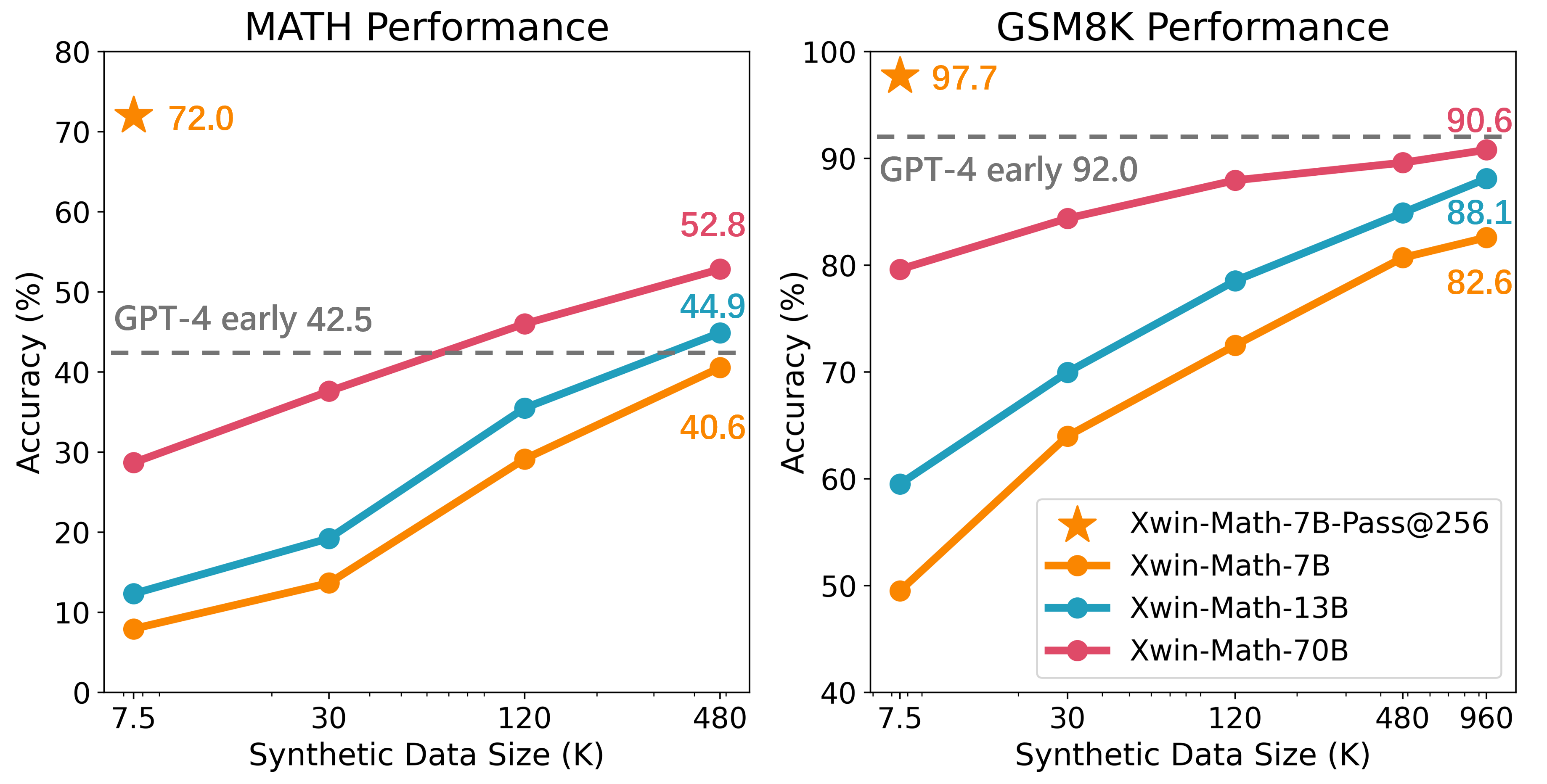}
    \caption{\label{fig:teaser}
    The orange star markers represent the accuracy achieved by selecting the best response from 256 random generations of the LLaMA-2 7B model. The high accuracy on the MATH (left) and GSM8K (right) benchmarks (72.0\% and 97.7\%, respectively) suggest that the LLaMA-2 7B already possesses strong mathematical capabilities, although the stability in generating correct answers could be enhanced. This paper demonstrates that by scaling synthetic SFT data, the stability can be significantly improved as evidenced by the curves. Through this straightforward scaling of SFT data, the top-performing model has exceeded an early GPT-4 model by 10.3\% on the MATH benchmark.
    }
    \vspace{-1.05em}
\end{figure}

\begin{table}[t]
\renewcommand\arraystretch{1.1}
\caption{Comparison of SFT data scaling with real versus synthetic math questions. It reveals that synthetic math questions are nearly as effective as real ones.}
\label{tab:within_7_5k}
\centering
\resizebox{.95\linewidth}{!}{
    \begin{tabular}{c|c|c|c|c}
    \toprule
    Data size & GSM8K-real & GSM8K-syn & MATH-real & MATH-syn \\ \midrule
    0.94K & 26.7 & 25.9 & 4.2 & 3.9 \\
    1.88K & 32.8 & 31.9 & 5.6 & 4.9 \\
    3.75K & 43.3 & 42.2 & 6.6 & 6.0 \\
    7.50K & 50.2 & 49.5 & 8.4 & 7.9 \\
    \bottomrule
    \end{tabular}
}
\end{table}

In this paper, we demonstrate that common language models of small size, such as the LLaMA-2 7B model \cite{touvron2023llama2}, already possess strong mathematical capabilities without specific pre-training on math-related data. Surprisingly, we find that with supervised fine-tuning on just thousands of math questions (noting that the SFT stage does not enhance capabilities as stated in \cite{bai2022training,ouyang2022training}), the model can correctly solve 97.7\% of GSM8K questions and 72.0\% of MATH questions, when selecting the best answer from 256 random generations, as indicated by the orange star marks in Figure~\ref{fig:teaser}. It is noteworthy that the accuracy has even outperformed those reported for the GPT-4 model, which achieved 92.0\% on GSM8K and 42.5\% on MATH \footnote{The accuracy numbers are reported in the GPT-4 technical report \cite{GPT4techreport}. GPT-4 models are continuously being improved. The latest GPT-4 Turbo (1106) API has increased accuracy to 94.8\% on GSM8K and 64.5\% on MATH. However, the LLaMA-2 7B model using the best of 256 generations still outperforms the latest GPT-4 models.}. Therefore, we conclude that the LLaMA-2 7B model has indeed developed strong mathematical capabilities. The primary issue is the lack of guarantee that the correct answer will be digged out, as most generations are incorrect. In fact, the accuracy drops to 49.5\% on GSM8K and 7.9\% on MATH if we consider only one random generation per question. We refer to this as the \emph{instability issue}.

To address the instability issue, we first observe that the accuracy improves almost in linear or even super-linear with exponentially increased supervised fine-tuning (SFT) data. Moreover, we note that the accuracy is far from reaching a plateau when utilizing all available GSM8K and MATH training data (as shown in Table ~\ref{tab:within_7_5k}). This observation encourages us to further scale up the SFT data. However, we face a challenge as there is a lack of publicly accessible real data to support this continuous scaling.

To overcome this limitation, we turn to synthetic data, employing a prestigious language model, namely GPT-4 Turbo, to produce synthetic math questions. We find that a straightforward ``brand-new'' generation strategy, which prompts the GPT-4 Turbo to create a completely new question based on preference ones and then applies a simple verifier (also GPT-4 Turbo based), has been highly effective. Specifically, as indicated in Table ~\ref{tab:within_7_5k}, the use of synthetically generated math questions can achieve accuracy nearly on par with that of real questions, highlighting the potential of synthetic SFT math questions for the scaling purpose.

Leveraging synthetic data has allowed us to scale our SFT data significantly, for instance, from 7.5K to 960K on GSM8K and from 7.5K to 480K on MATH. This data scaling shows nearly perfect scaling behavior, as drawn in Figure~\ref{fig:teaser}. Specifically, by simply scaling the SFT data, our model has become the first to exceed 80\% and 40\% accuracy on GSM8K and MATH, respectively, using a standard LLaMA-2 7B base model (achieving 82.6\% and 40.6\% respectively)\footnote{Concurrently, DeepSeek-MATH-7B~\cite{shao2024deepseekmath} also surpasses 80\% accuracy. However, their approach relies on a much stronger base model extensively pre-trained on math-related corpora and a sophisticated RL algorithm. Our results are complementary to theirs.}.

The straightforward synthetic SFT data proves effective from stronger base models as well, such as LLaMA-2 70B, which achieves 90.6\% on GSM8K and 52.8\% on MATH. To the best of our knowledge, this is the first open-source model to exceed 90\% accuracy on GSM8K. It is also the first open-source model to outperform GPT-4 (i.e., GPT-4-0314) on the MATH benchmark, demonstrating the efficacy of our simple synthetic scaling method.

In addition to the strong results, we have also gleaned insights into the effectiveness of our approach: 1) As the scale of SFT data increases, the model's accuracy tends to plateau when utilizing 256 attempts; however, there is a marked increase using 1 response. This indicates that while the model's upper capability limit remains fairly constant, the performance gains are primarily due to enhanced stability in generating correct answers. 2) The accuracy of solving math problems follows a power law with respect to the number of chain-of-thought (CoT) steps with different SFT data quantities. An expanded SFT dataset improves the reliability of each reasoning step. Further increasing the proportion of training samples with longer CoT steps through resampling can significantly improve the accuracy of the model for difficult questions. 3) An analysis of error types during the scaling process reveals that calculation errors are more readily mitigated compared to reasoning errors.

\section{Examine Math Capability of Language Models}
\label{sec:motivation}

\paragraph{Metrics}

We employ two metrics to examine the math capabilities of language models.

The first is a Pass@N metric
\begin{equation}
    \text{Pass@N} = \mathop{\mathbb{E}}_{\text{Problems}} \left[ \min(c, 1) \right],
\end{equation}
where $c$ represents the number of correct answers out of $N$ responses. This metric considers a question to be solved if at least one correct answer is produced from $N$ random generations. We employ this metric to reflect the potential or capability of a model in solving a math question. To enhance the diversity of the $N$ generations, we set the temperature of the generation process to 0.7\footnote{It is worth noting that most math models utilize a greedy generation strategy with the temperature set to 0. However, the impact of this difference is minimal.}.

The second is a PassRatio@N metric
\begin{equation}
    \text{PassRatio@N} = \mathop{\mathbb{E}}_{\text{Problems}} \left[ \frac{c}{N} \right],
\end{equation}
which measures the percentage of correct answers within the $N$ generated answers. This metric is somewhat equivalent to Pass@1, but with reduced variance.

\paragraph{Observations}
Based on these two metrics, we examine the performance of the LLaMA-2 models on the GSM8K and the MATH benchmarks\footnote{Following \cite{lightman2023let}, we utilize a subset of 500 test samples from the MATH benchmark for experimental efficiency.} as shown in Figure~\ref{fig:teaser}. To adapt models for these two benchmarks in instruction-following settings, we use their SFT versions, which are trained with a limited amount of SFT data (i.e., 7.5K). As demonstrated in \cite{bai2022training,ouyang2022training}, the SFT stage does not enhance capabilities (and may even lead to a reduction, as mentioned in the context of ``alignment taxes''). Therefore, employing the SFT version provides a fair assessment of the models' mathematical capabilities.

We first observe that the Pass@256 metrics for the LLaMA-2 7B model on both benchmarks are remarkably high: 97.7\% on GSM8K and 72.0\% on MATH. This suggests that the LLaMA-2 7B model possesses a strong capability for solving mathematical problems.

We then notice that the PassRatio@256 is significantly lower than that of Pass@256, being 48.2\% on GSM8K and 7.9\% on MATH. This suggests that while the correct answers to most math questions are present within 256 random generations, there is no assurance that the correct answers will consistently be extracted, a phenomenon we refer to as an "instability issue".

In the following, we will present a simple approach to significantly reduce the \emph{instability issue}.

\section{Scaling SFT Data using Synthetic Math Questions}
In this section, we first demonstrate that scaling up the limited real SFT data can significantly alleviate the instability issue. We also observe that the accuracy has not yet plateaued when using the full available GSM8K and MATH training data. We consider further scaling up SFT data using synthetic math questions. To this aim, we introduce a straight-forward method for synthetic data generation utilizing the GPT-4 Turbo API. The synthetic data proves to be as effective as real math questions. Consequently, we boldly scale the synthetic SFT data to 960K on GSM8K and 480K on MATH, respectively, resulting in nearly perfect scaling behavior, and reach state-of-the-art accuracy.

\paragraph{Scaling using Real Math Questions} We begin by examining the scaling behavior of real math questions across the entire GSM8K and MATH training sets. 
As indicated in Table ~\ref{tab:within_7_5k}, we observe a consistent accuracy improvement, increasing from 26.7\% to 50.2\% on GSM8K, and from 4.2\% to 8.4\% on MATH, with no signs of saturation.

\paragraph{Synthetic SFT Data Generation} Since the real data has been exhausted, we contemplate further scaling up SFT data using synthetically generated math questions.

We introduce a straightforward three-step approach with the assistance of the GPT-4 Turbo API:
\begin{itemize}
\item {\emph{Step 1. Generate a new math question.} 
We request the GPT-4 Turbo API to generate a brand-new question using a reference math question as a starting point. To improve the validity of the new questions, we incorporate three rules into the prompt: Firstly, the new question must obey common knowledge; secondly, it should be solvable independently of the original question; and thirdly, it must not include any answer responses. Besides, we have set specific formatting requirements for questions and answers tailored to various target datasets.}
\item {\emph{Step 2. Verify the question.} We further enhance the quality of the generated questions by validating and refining them through attempted solutions. By integrating solving and verification steps into a single prompt, we have found that this approach consistently elevates the validity of questions across different benchmarks.}
\item {\emph{Step 3. Generate chain-of-thought (CoT) answers.} We request GPT-4 Turbo to produce a chain-of-thought (CoT) answer response for each newly generated question.}
\end{itemize}

The detailed prompt designs are shown in Appendix~\ref{sec:appendix_prompt}.

\paragraph{Comparison of Synthetic SFT Data versus Real Data} To assess the quality of the synthetically generated math questions, we evaluate their effectiveness against real questions from the GSM8K and MATH training sets, utilizing a LLaMA-2 7B model, as detailed in Table ~\ref{tab:within_7_5k}. The results indicate that the synthetic math questions are nearly as effective as the real ones.

We also explored various other synthetic methods as proposed in previous works~\cite{xu2023wizardlm, yu2023metamath, an2023learning}. These methods also prove to be effective, though marginally less so than the our approach, as illustrated in Figure~\ref{fig:prompt}.

\paragraph{Scaling to about a Million SFT Math Data} 
Considering the effectiveness of the synthetic approach, we substantially increase the scale of the SFT data for both GSM8K and MATH problems, to 960K and 480K, respectively. Figure~\ref{fig:teaser} presents the main reasults utilizing various sizes of the LLaMA-2 series. The straightforward scaling strategy yields state-of-the-art accuracy. 

It is also worth noting that the accuracy has not yet reached its peak. Exploring the effects of additional scaling will be left as our future research.

\section{Experiments}
\subsection{Datasets and Evaluations}

We conduct experiments on 5 benchmarks to evaluate the efficacy of the proposed method. 

\noindent \textbf{GSM8K}~\cite{cobbe2021training}. This is a high-quality, linguistically diverse math dataset, whose math knowledge mainly covers grade school level. It includes 7,473 training examples and 1,319 test cases. In this work, we use its training set as the given questions to generate new synthetic data.  

\noindent \textbf{MATH}~\cite{hendrycks2021measuring}. 
This dataset focuses on competitive-level math problems that requires high levels of reasoning ability and mathematical knowledge. It consists of 7,500 training examples and 5,000 test cases. We use the training examples to generate synthetic data.

\noindent \textbf{SVAMP}~\cite{patel2021nlp}.
This dataset comprises elementary-level math problems. We utilize all 1,000 of its test cases to assess the cross-dataset performance of our models.

\noindent \textbf{ASDiv}~\cite{miao2021diverse}. This dataset contains a set of math problems with diverse language patterns and types of questions. We adopt the test set of 2,305 problems as evaluation benchmark. 

\noindent \textbf{Hungarian National High School Exam} This evaluation benchmark is first introduced by Grok-1~\cite{grok1}, which is designed for evaluating the out-of-domain capability of math models. It consists of 33 challenging problems. 

It is worth noting that the final answers of Hungarian National High School Exam dataset is annotated by human, while other benchmarks are labelled using automatic scripts, similar to previous works 
 \cite{luo2023wizardmath, gou2023tora}.

\subsection{Implementation Details}
In data synthesis, we utilize the GPT-4 Turbo API, setting the temperature to 1.0 for both question and answer generation. 

For supervised fine-tuning, we employ the Adam optimizer with a cosine learning rate schedule spanning a total of 3 epochs of training. The maximum learning rate is set 2e-5 (except that 2e-6 for the Mistral-7b model) and there is a 4\% linear warm-up. The maximum token length is set 2048, and the Vicuna-v1.1~\cite{zheng2023judging} system prompt is used. All experiments are conducted on 8$\times$Nvidia H100 GPUs. Our most resource-intensive experiment, involving a 70B model and 960K data points, takes 1900 H100 GPU hours.

For evaluation, we use the same prompt as used in SFT and set the maximum sequence length to 2048. The vLLM~\cite{kwon2023efficient} is used in answer generation.

\begin{table}[h]
\renewcommand\arraystretch{1.1}
\caption{Math reasoning performances of various LLMs.
}
\label{tab:main_table}
\centering
\resizebox{.95\linewidth}{!}{
\begin{tabular}{@{}lcc@{}}
\toprule
\multicolumn{1}{c}{Model} & GSM8K & MATH \\ \midrule \midrule
\multicolumn{3}{c}{\textit{Closed-source models}} \\
GPT-4 Turbo (1106) & 94.8 & 64.5 \\
GPT-4-0314 & 94.7 & 52.6 \\
GPT-4~\cite{achiam2023gpt} & 92.0 & 42.5 \\
Claude-2~\cite{claude2} & 88.0 & - \\
GPT-3.5-Turbo~\cite{gpt35turbo} & 80.8 & 34.1 \\
\midrule
\multicolumn{3}{c}{\textit{Open-source models LLaMA-2-7B}} \\
WizardMath-7B~\cite{luo2023wizardmath} & 54.9 & 10.7 \\
MuggleMath-7B~\cite{li2023query} & 68.4 & - \\
MetaMath-7B~\cite{yu2023metamath} & 66.5 & 19.8 \\
LEMA-LLaMA-2-7B~\cite{an2023learning} & 54.1 & 9.4 \\
\cellcolor{gray!25}Xwin-Math-7B (ours) & \cellcolor{gray!25}\textbf{82.6} & \cellcolor{gray!25}\textbf{40.6} \\ \midrule
\multicolumn{3}{c}{\textit{Open-source models Mistral-7B}} \\
WizardMath-7B-v1.1~\cite{luo2023wizardmath} & 83.2 & 33.0 \\
MetaMath-Mistral-7B~\cite{yu2023metamath} & 77.4 & 28.2 \\
\cellcolor{gray!25}Xwin-Math-Mistral-7B (ours) & \cellcolor{gray!25}\textbf{89.2} & \cellcolor{gray!25}\textbf{43.7} \\ \midrule
\multicolumn{3}{c}{\textit{Open-source models Llemma-7B}} \\
MetaMath-Llemma-7B~\cite{yu2023metamath} & 69.2 & 30.0 \\
\cellcolor{gray!25}Xwin-Math-Llemma-7B (ours) & \cellcolor{gray!25}\textbf{84.2} & \cellcolor{gray!25}\textbf{47.2} \\ \midrule
\multicolumn{3}{c}{\textit{Open-source models LLaMA-2-13B}} \\
WizardMath-13B~\cite{luo2023wizardmath} & 63.9 & 14.0 \\
MuggleMath-13B~\cite{li2023query} & 74.0 & - \\
MetaMath-13B~\cite{yu2023metamath} & 72.3 & 22.4 \\
LEMA-LLaMA-2-13B~\cite{an2023learning} & 65.7 & 12.6 \\
\cellcolor{gray!25}Xwin-Math-13B (ours) & \cellcolor{gray!25}\textbf{88.1} & \cellcolor{gray!25}\textbf{44.9} \\ \midrule
\multicolumn{3}{c}{\textit{Open-source models LLaMA-2-70B}} \\
WizardMath-70B~\cite{luo2023wizardmath} & 81.6 & 22.7 \\
MuggleMath-70B~\cite{li2023query} & 82.3 & - \\
MetaMath-70B~\cite{yu2023metamath} & 82.3 & 26.6 \\
LEMA-LLaMA-2-70B~\cite{an2023learning} & 83.5 & 25.0 \\
\cellcolor{gray!25}Xwin-Math-70B (ours) & \cellcolor{gray!25}\textbf{90.6} & \cellcolor{gray!25}\textbf{52.8} \\ \bottomrule
\end{tabular}
}
\vspace{-1em}
\end{table}

\subsection{Main Results and Comparison with State-of-the-art Models}

In this comparison, we examine both in-domain benchmarks, GSM8K/MATH, and out-of-domain benchmarks, such as the Hungarian National High School Exam. For in-domain evaluation of each benchmark, we utilize data synthesized from its respective training samples. For GSM8K, 960K synthetic data is employed, while for MATH, 480K synthetic data is used. For out-domain evaluation, we test models trained using GSM8K, MATH, or a mixed of two synthetic sets.

For base models, we consider both common language models, i.e., LLaMA-2 7B/13B/70B/Mistral-7B, and math-specific models, such as Llemma-7B, to assess the generality of the proposed approach.

\paragraph{In-Domain Results}

Table ~\ref{tab:main_table} presents a comparison of the proposed approach with the state-of-the-art open and closed-source models. Across all base models, our method significantly outperforms the previous best approaches that use the same pre-trained base model. 

On LLaMA-2-7B, our approach exceeds the prior best by absolutely +14.2 on GSM8K (compared to MuggleMath-7B~\cite{li2023query}), and by +20.8 on MATH (compared to MetaMath-7B~\cite{yu2023metamath}), respectively. It even surpasses several latest 70B models dedicated for math capabilities, such as WizardMath-70B \citep{luo2023wizardmath} (82.6 versus 81.6 on GSM8K). On LLaMA-2-13B, the improvements are +14.1 on GSM8K (compared to MuggleMath-13B~\cite{li2023query}) and +22.5 on MATH (compared to MetaMath-13B~\cite{yu2023metamath}), respectively. On LLaMA-2-70B, the gains are +7.1 on GSM8K (compared to LEMA-LLaMA-2-70B~\cite{an2023learning}) and +26.2 on MATH (compared to MetaMath-70B~\cite{yu2023metamath}), respectively.

On a stronger common language model, i.e., Mistral-7B, the improvements are +6.0 on GSM8K and +10.7 on MATH (compared to WizardMath-7B-v1.1~\cite{luo2023wizardmath}), respectively.

On a math-specific base model, such as Llemma-7B, the gains are +15.0 on GSM8K and +17.2 on MATH (compared to MetaMath-Llemma-7B~\cite{luo2023wizardmath}), respectively.

It is also noteworthy that our LLaMA-2-70B model achieves competitive accuracy with early versions of GPT-4 on GSM8K and MATH. To our knowledge, this is the first LLaMA-based model to outperform GPT-4-0314 on MATH.

These results demonstrate the significant effectiveness and broad applicability of scaling synthetic math SFT data.

\paragraph{Out-of-Domain Results}

We test the models trained using GSM8K, MATH, or a mixed of two synthetic sets on an out-of-domain benchmark, Hungarian National High-School Exam Test, following the practice in~\cite{grok1}. 

Table ~\ref{tab:Hungarian} shows the results. Our model trained on the mixing data (240K MATH synthetic data + 240K GSM8K synthetic data) ranked as the second, just behind the GPT-4 and much better than other models. Additionally, we plot the correlation between GSM8K and Hungarian national high-school exam scores in Appendix~\ref{sec:appendix_results}. The results show that there is no significant benchmark overfitting in our model.

\begin{table}[t]
\renewcommand\arraystretch{1.1}
\caption{Hungarian national high school exam test result of various LLMs.
}
\label{tab:Hungarian}
\centering
\resizebox{.9\linewidth}{!}{
\begin{tabular}{l|c}
\toprule
Model & Test Score (\%) \\ \midrule
GPT-4~\cite{achiam2023gpt} & 68 \\
Grok-1~\cite{grok1} & 59 \\
Claude-2~\cite{claude2} & 55 \\
GPT-3.5 Turbo~\cite{gpt35turbo} & 41 \\
DeepSeek-LLM-67B-Chat~\cite{bi2024deepseek} &  58 \\
\midrule
Xwin-Math-70B (480K GSM8K) & 22 \\
Xwin-Math-70B (120K MATH) & 51 \\
Xwin-Math-70B (480K MATH) & 59 \\
Xwin-Math-70B (480K Mix) & 65 \\
\bottomrule
\end{tabular}
}
\end{table}

\begin{figure*}[ht]
\vspace{-1em}
    \centering
    \includegraphics[width=0.95\textwidth]{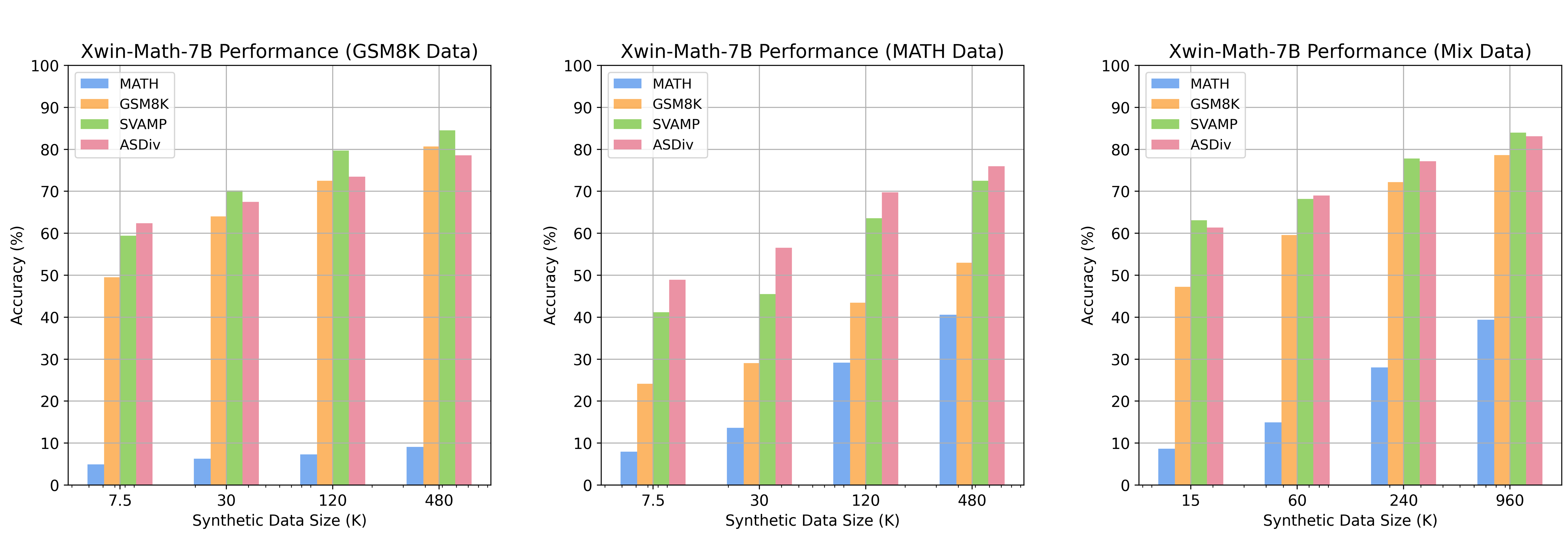}
    \caption{\label{fig:mix_dataset}
    Comparing the increase in SFT data scale using either a single dataset or mixed datasets.
    }
\end{figure*}

Figure.~\ref{fig:mix_dataset} (Left) presents the results of the model trained on GSM8K synthetic data, while Figure.~\ref{fig:mix_dataset} (Middle) presents the results of the model trained on MATH. We find that the accuracy of other benchmarks also improves as the amount of data increases for models trained with either GSM8K or MATH synthetic data. We also note that the generalization behaviors differ for GSM8K and MATH models: 1) SVAMP and ASDiv benefit more from GSM8K models than from MATH models. 2) While MATH models perform relatively well on the GSM8K benchmark, GSM8K models perform considerably worse on MATH benchmarks.

Figure.~\ref{fig:mix_dataset} (Right) shows the results of models using a mixture of GSM8K and MATH in a 1:1 ratio. These models exhibit balanced scaling behaviors in both in-domain and out-of-domain benchmarks.

\begin{figure*}
    \centering
    \includegraphics[width=0.75\textwidth]{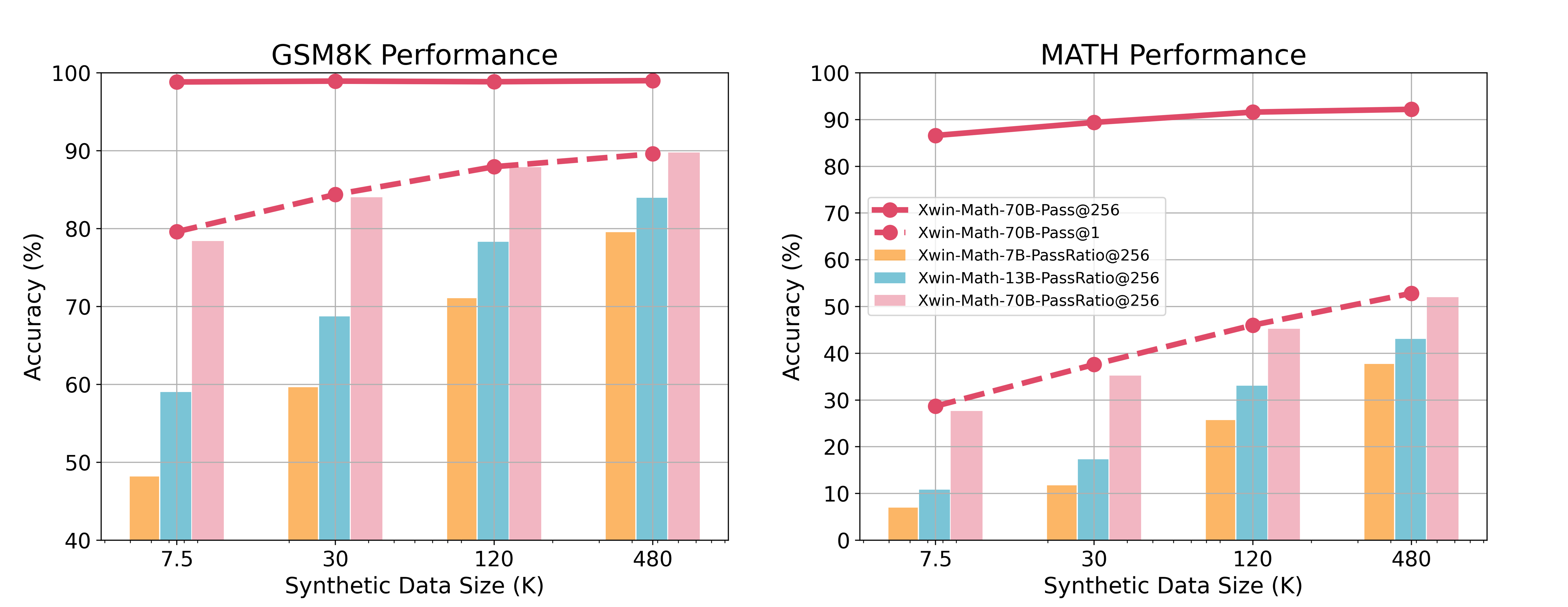}
    \caption{\label{fig:ratio}
    The Pass@256 and PassRatio@256 curve  with increasing data size on GSM8K and MATH benchmark.
    }
    \vspace{-1em}
\end{figure*}

\subsection{What Happens behind Performance Improvements?}
\noindent \textbf{Pass@256 \textit{v.s.} PassRatio@256} 
To deepen the understanding behind the performance improvements, we tracked Pass@N metric and PassRatio@N metric under different data size. The results are shown in Figure \ref{fig:ratio}. With very limited synthetic data (\textit{e.g.} 7.5K samples), the Xwin-Math-70B model already has very high Pass@256, indicating the strong ability to generate correct answers through multiple attempts. Meanwhile, the Pass@256 metric only changed slightly with increasing the amount of used data. In contrast, PassRatio@256, which reflects the stability to generate correct answer, increases significantly with the amount of synthetic data, and its growth trend is similar to that of Pass@1. This result confirms our hypothesize that the performance improvements is mainly caused by the better stability in answer generation rather than stronger ability to answer the question.

\noindent \textbf{Estimated Single-step Reasoning Accuracy}  
Because of the Chain-of-Thought (CoT) are adopted in inference, the process of answer mathematical problems is completed by a multi-step reasoning process. Therefore, we hypothesize that the increase in final answer accuracy can be interpreted by the improvement in single-step reasoning accuracy. Based on this assumption, if one question can be theoretically answered by $s$ reasoning steps in CoT, the final answer accuracy can be approximate by the power function of the single-step reasoning accuracy:
\begin{equation}
\text{Acc}_{\text{final}} = \text{Acc}_{\text{step}}^s
\label{eq:step_acc}
\end{equation}

\begin{figure}
    \centering
    \includegraphics[width=0.48\textwidth]{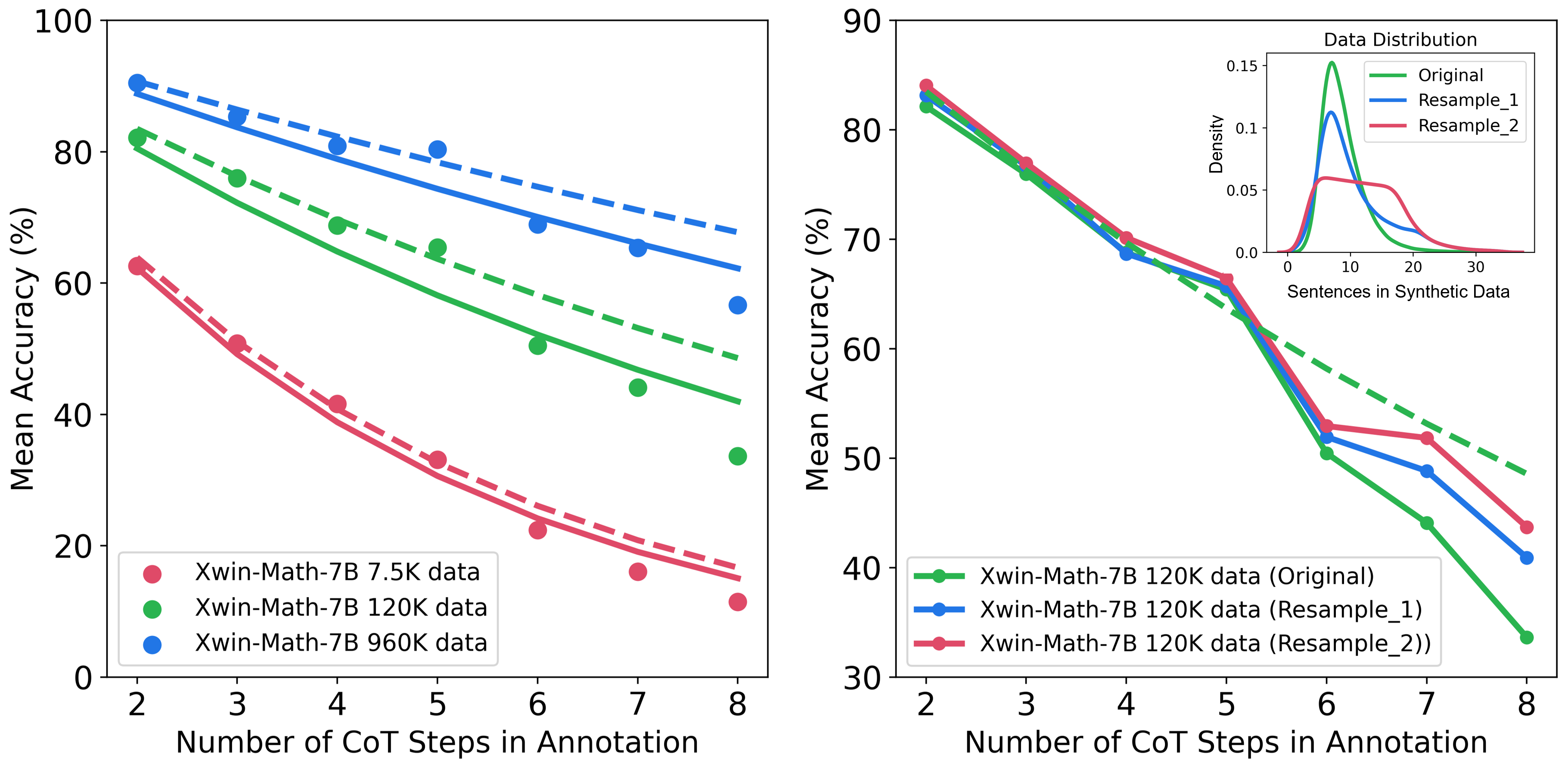}
    \caption{\label{fig:step_analysis}
    Left: The relationship between the mean accuracy on the GSM8K and the number of annotated CoT steps with data increasing. The solid line is fitted using all seven points, while the dashed line is fitted using the first four points. Right: Changes in mean accuracy when resampling is used to increase the CoT lengeh of training data.
    }
\end{figure}

With this equation, step accuracy can be estimated from the final answer accuracy. We experimented on GSM8K. For each question in the test set, we generated 256 responses and used the number of steps in the GSM8k test set's CoT annotations as the theoretical CoT steps. We draw the curve to show the relationship between the number of CoT reasoning steps and mean final answer accuracy and show the fitted curve based on Equation.~\ref{eq:step_acc}. We test Xwin-Math-7B models with different synthetic data, and the results are shown in Figure \ref{fig:step_analysis}. The solid line is fitted using all seven points and Table ~\ref{tab:single-step-reasoning} shows the estimated single-step accuracy when using different amounts of data using all data points, and it can be seen that the single-step accuracy improve significantly with more data.

However, when we fit based on Equation.~\ref{eq:step_acc} to the first four points, as shown in dashed lines, we found that the latter three points were significantly below the curve. We believe this phenomenon may be related to the smaller proportion of more complex problems in the training data. Therefore, we resampled the 960K synthetic data according to the number of sentences in CoT solution. As can be seen from Figure~\ref{fig:step_analysis} (right), when the proportion of complex problems is increased, the accuracy for simpler problems remains virtually unchanged, but the accuracy for more complex problems can be significantly improved. Moreover, the utilization of data resampling can increase the model's PassRatio@256 from 71.1 to 72.8. This experimental result provides new insights into data selection for mathematical reasoning tasks.

In addition, we further used the GPT-4 Turbo to find the position where the first step in our answer was wrong and normalized that position by the total number of steps in each answer. As the estimated single-step accuracy gets higher, the first error position of the normalization is postponed.

\begin{table}[t]
\renewcommand\arraystretch{1.1}
\caption{The estimated single-step reasoning accuracy and the average normalized first error position by GPT-4 Turbo in Xwin-Math-7B on GSM8K benchmark.}
\label{tab:single-step-reasoning}
\centering
\resizebox{.95\linewidth}{!}{
    \begin{tabular}{l|c|c}
    \toprule
    Data size & Estimated $\text{Acc}_{\text{step}}$ & Normalized first error position \\ \midrule
    7.5K & 78.9 & 67.1 \\
    120K & 89.7 & 83.9 \\
    960K & 94.2 & 90.9 \\
    \bottomrule
    \end{tabular}
}
\end{table}

\begin{figure}
    \centering
    \includegraphics[width=0.4\textwidth]{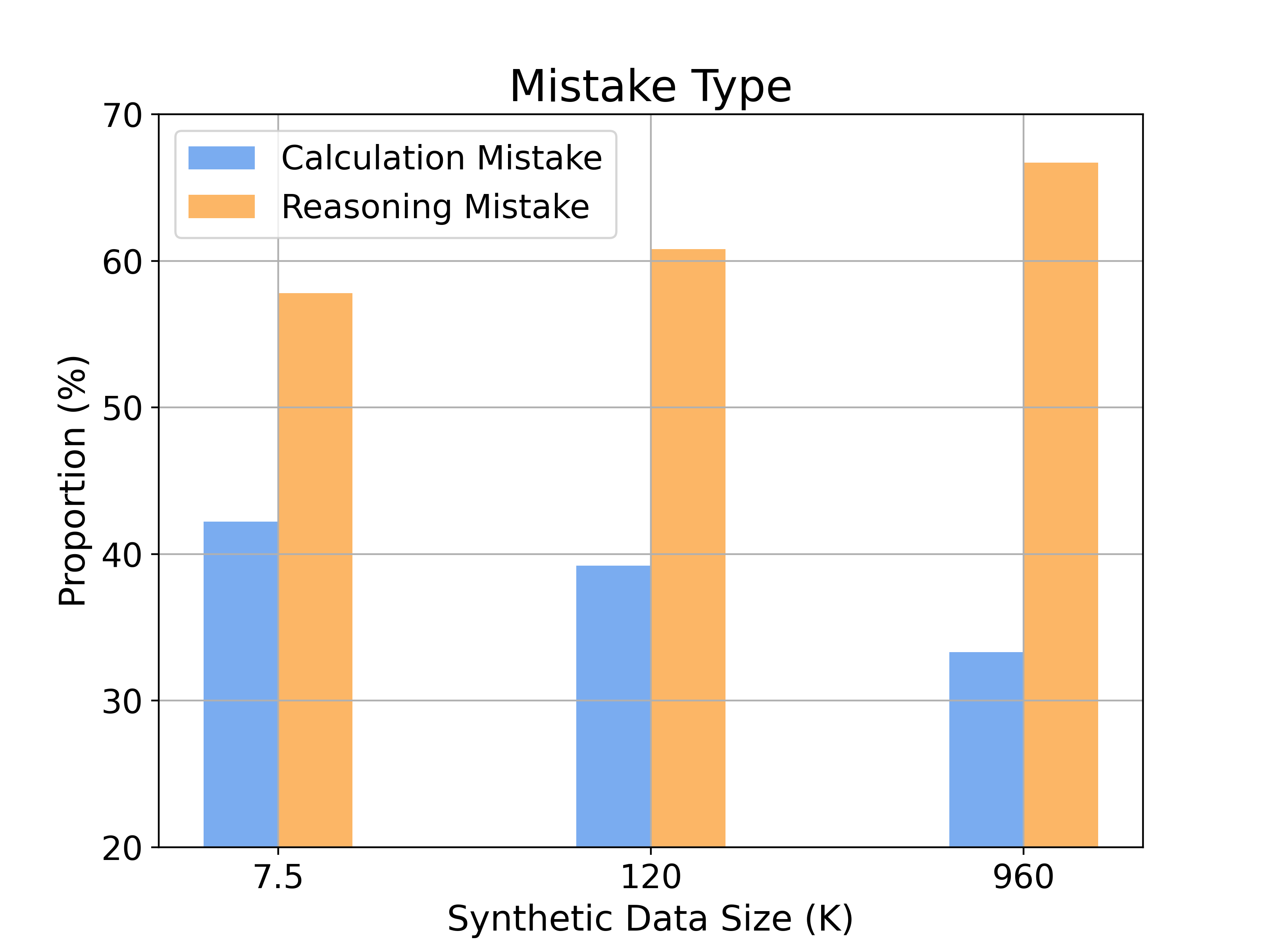}
    \caption{\label{fig:mistake_type}
    Changes in the proportion of calculation and reasoning mistake during data increased.
    }
    \vspace{-1em}
\end{figure}

\noindent \textbf{The Improvement in the Accuracy of Numerical Calculations is More Significant than Logical Reasoning}
The performance of the model gradually improves as the synthetic data increases. For a deeper understanding, we analyze the error proportion for different types of errors on GSM8K. 
We categorized errors into two types: reasoning errors and calculation errors. Reasoning errors primarily encompass issues such as loss of conditions and concept confusion, while calculation errors include incorrect analysis of quantitative relationships and numerical computation mistakes. Based on the experimental results illustrated in Figure \ref{fig:mistake_type}, we observe a gradual decrease in the percentage of calculation errors, suggesting that GSM8K is correcting calculation errors at a faster rate than reasoning errors.

\subsection{Ablations on the Data Synthetic Schema}
\noindent \textbf{Comparison with Other Data Synthetic Methods}
We compared our approach with following common used data synthetic methods:

\noindent \textit{Add Constraint.} Adding one more constrain to the original question while keeping others unchanged, which is used in WizardMath and MuggleMath.

\noindent \textit{Change Numbers.} Changing the numbers that appear in the problem while keeping the context intact. which is used in MuggleMath. 

\noindent \textit{Change Background.} Changing the background in the question while keeping others the same. 

\noindent \textit{The Combination of Changing Numbers and Background.} A hybrid approach that combines changing both numbers and background.

\noindent \textit{MetaMath Approach.} The synthetic methods proposed in MetaMath, including answer augmentation, rephrasing question, self-verification question and FOBAR question. In experiments, we follow the implementation of MetaMath but use GPT-4 Turbo instead of GPT-3.5 Turbo to generate response data using their released questions.

\begin{figure}
\vspace{-2em}
    \centering
    \includegraphics[width=0.45\textwidth]{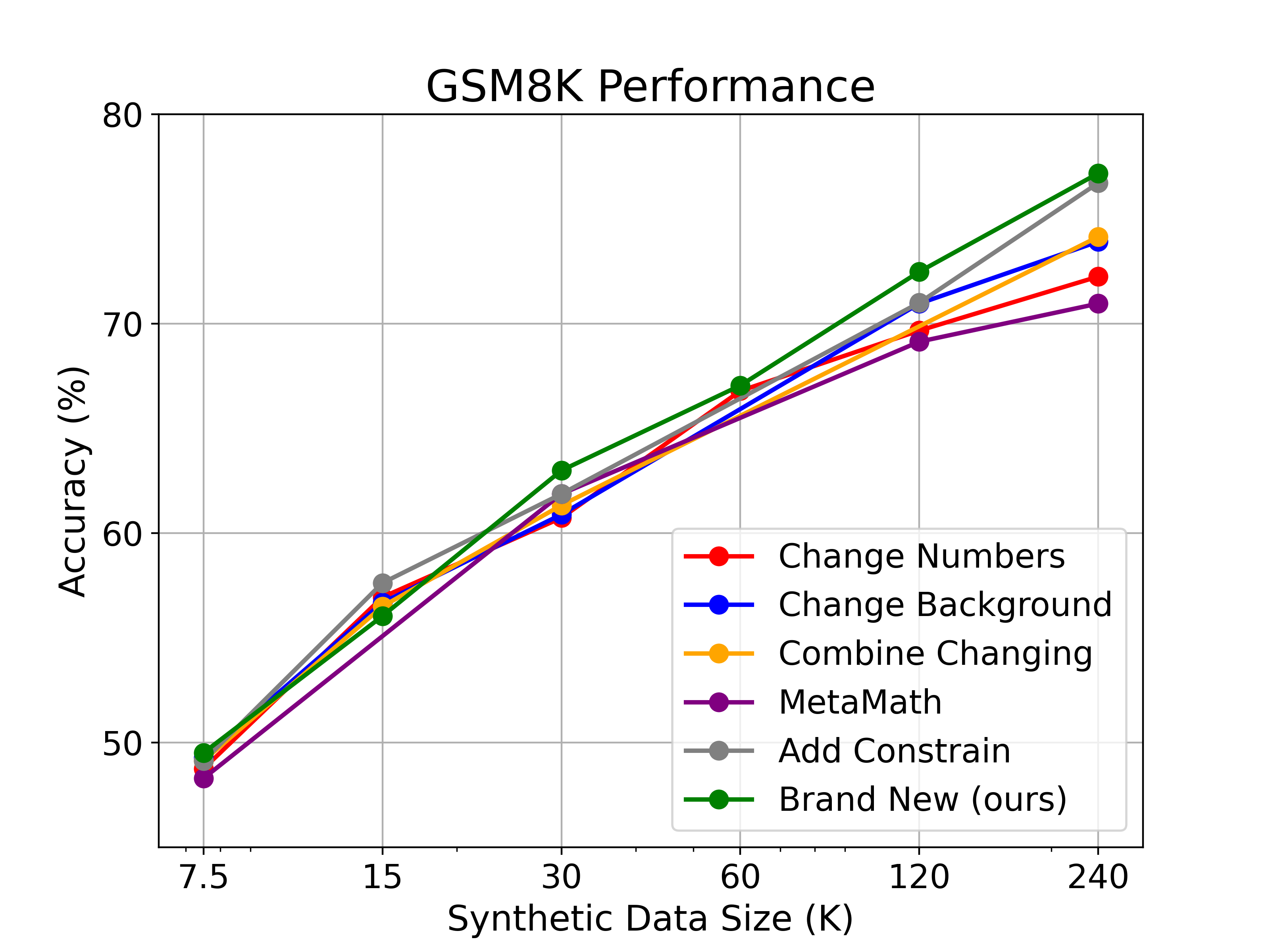}
    \caption{\label{fig:prompt}
    GSM8K and MATH performance of different synthetic methods.
    }
\vspace{-1em}
\end{figure}

The experimental results in the Figure \ref{fig:prompt} show that when the data size is relatively small, \textit{e.g.}, 7.5k and 30k samples, the performance gap between the different methods is negligible. However, as the data size increases, our method and the method with added constraints show stronger performance. This suggests that the choice of data synthetic strategy becomes more critical as the data size increases, and that some methods can scale the data more efficiently, thus improving the performance.

\begin{table}[t]
\renewcommand\arraystretch{1.1}
\caption{Ablation of question verification on MATH.}
\label{tab:question-verification}
\centering
\resizebox{.99\linewidth}{!}{
\begin{tabular}{l|c}
\toprule
Model & Pass@1 (\%) \\ \midrule
Xwin-Math-70B (7.5K data) & 28.9 \\
Xwin-Math-70B (7.5K data) w/o verification & 28.1 (-0.8) \\
Xwin-Math-70B (30K data) & 37.6 \\
Xwin-Math-70B (30K data) w/o verification & 36.6 (-1.0) \\
\bottomrule
\end{tabular}
}
\vspace{-1.5em}
\end{table}

\noindent \textbf{Effects of Question Verification.}
The question verification is used to further improve the generation quality. In our experiments, we found it can improve the performance on MATH benchmark, the results are shown in Table ~\ref{tab:question-verification}, while we do not see significantly impact on GSM8K dataset.

\section{Related Works}
\paragraph{Large Language Models}
Large language models~\cite{brown2020language,achiam2023gpt,touvron2023llama,touvron2023llama2} have made significant achievements, with impressive performance on a wide range of tasks. Currently, closed-source large language models, represented by GPT~\cite{brown2020language,achiam2023gpt}, Gemini~\cite{team2023gemini}, Grok~\cite{grok1}, and Claude-2~\cite{claude2}, are the most advanced models in terms of performance. However, open-source models, represented by LLaMA~\cite{touvron2023llama}, LLaMA-2~\cite{touvron2023llama2} and Mixtral~\cite{jiang2024mixtral}, have also progressed rapidly, and have even shown competitive performance with the closed-source models on some tasks. Our work, which aims to improve the performance of open-source LLMs on mathematical tasks by fine-tuning them on synthetic data.

\paragraph{Reasoning Framework for Improving Mathematical Capability}
Chain-of-thoughts~\cite{wei2022chain}  encourages the LLMs perform multi-step reasoning by specific designed prompts and can improve reasoning performance. Based on this work, many subsequent works suggesting further improvements~\cite{fu2022complexity, zhang2022automatic, kojima2022large}. The above works focus primarily on how to improve performance through better prompt design or inference strategies without fine-tuning the model, whereas our work focuses on how to improve the model itself, and thus these approaches are complementary to ours.

\paragraph{Fine-tuned LLM for Math}
Another sort of works~\cite{lightman2023let,luo2023wizardmath,azerbayev2023llemma,yue2023mammoth,yu2023metamath,an2023learning,li2023query, gou2023tora} try to improve performance directly by training the model on mathematical data. 
A direct way is to use fine-tuning to improve models. One widely used method is to use synthetic data, which is very close to our approach: MetaMath~\cite{yu2023metamath} presents to bootstrap questions to augment data. LeMA~\cite{an2023learning} collects mistake-correction data pairs by using GPT-4 as a corrector. And MuggleMath~\cite{li2023query} augments the GSM8K dataset by incorporating GPT-4 with a series of pre-defined operations. Compared to these synthetic data based efforts, our data synthetic method is much simpler and more scalable due to introduce less prior and constraint. 

\paragraph{SFT Data Scaling}
Recently, some research efforts have focused on the data scale for supervised fine-tuning. For instance, LIMA~\cite{zhou2023lima} mentions that fine-tuning with 1,000 high-quality instructions can yield impressive results in various general tasks. Other studies have indicated that performance scales with data size in mathematical and coding tasks~\cite{dong2023abilities}. Recent work~\cite{bi2024deepseek} even uses 1.5 million data for instruct fine-tuning to obtain top performance. However, the intrinsic reasons behind this scaling effect have not been thoroughly investigated.

\vspace{-0.5em}
\section{Conclusion}
\vspace{-0.5em}

This study reveals that common 7B language models, such as LLaMA-2 7B, already exhibit strong mathematical capabilities, challenging the previous belief that advanced mathematical reasoning is exclusive to larger, more extensively pre-trained models. By significantly scaling up SFT data, we have markedly improved the stability of the model's mathematical problem-solving skills. Our methodology has enabled the Xwin-Math models to reach performance levels comparable to, and in some instances surpassing, those of their larger counterparts. Our analysis also indicates that the enhancements are primarily attributable to heightened accuracy in single-step reasoning and a extra resampling of training data can improve the accuracy of harder questions. Additionally, we see more substantial reduction of calculation errors as opposed to logical reasoning errors. Our research contributes valuable insights into the mathematical capabilities of large language models.

\section*{Acknowledgments}
Chen Li and Nanning Zheng were supported in part by NSFC under grant No. 62088102. Thank Shengnan An at IAIR, Xi’an Jiaotong University for his valuable advice on this work.

\bibliography{custom}

\appendix

\clearpage\onecolumn
\section{Synthetic Prompt on GSM8K}
\label{sec:appendix_prompt}

\begin{user_example}[frametitle={Prompt 1: Question Generation}]
{
    \fontsize{10pt}{9pt}\selectfont Please act as a professional math teacher. 
    
    Your goal is to create high quality math word problems to help students learn math.
    
    You will be given a math question. Please create a new question based on the Given Question and following instructions.
    
    To achieve the goal, you have three jobs.
    
    \# Please generate a similar but new question according to the Given Question.
    
    \# Check the question by solving it step-by-step to find out if it adheres to all principles.
    
    \# Modify the created question according to your checking comment to ensure it is of high quality.
    
    You have five principles to do this.
    
    \# Ensure the new question only asks for one thing, be reasonable, be based on the Given Question, and can be answered with only a number (float or integer). For example, DO NOT ask, ‘what is the amount of A, B and C?’. 
    
    \# Ensure the new question is in line with common sense of life. For example, the amount someone has or pays must be a positive number, and the number of people must be an integer.
    
    \# Ensure your student can answer the new question without the given question. If you want to use some numbers, conditions or background in the given question, please restate them to ensure no information is omitted in your new question.
    
    \# Please DO NOT include solution in your question.
    
    \# If the created question already follows these principles upon your verification. Just keep it without any modification.
    
    Given Question: {given question}
    
    Your output should be in the following format:
    
    CREATED QUESTION: <your created question>
    
    VERIFICATION AND MODIFICATION: <solve the question step-by-step and modify it to follow all principles>

    FINAL CREATED QUESTION: <your final created question>
}
\end{user_example}

\begin{user_example}[frametitle={Prompt 2: Answer Generation}]
{
    \fontsize{10pt}{9pt}\selectfont Please act as a professional math teacher.
    
    Your goal is to accurately solve a math word problem.
    
    To achieve the goal, you have two jobs.
    
    \# Write detailed solution to a Given Question.
    
    \# Write the final answer to this question.
    
    You have two principles to do this.
    
    \# Ensure the solution is step-by-step.
    
    \# Ensure the final answer is just a number (float or integer).
    
    Given Question: {given question}
    
    Your output should be in the following format:
    
    SOLUTION: <your detailed solution to the given question>
    
    FINAL ANSWER: <your final answer to the question with only an integer or float number>
}  
\end{user_example}

\begin{user_example}[frametitle={Prompt 3: Question Generation w/o verification}]
{
    \fontsize{10pt}{9pt}\selectfont Please act as a professional math teacher.
    
    Your goal is to create high quality math word problems to help students learn math.
    
    You will be given a math question. Please create a new question based on the Given Question and following instructions.
    
    To achieve the goal, you have one job.
    
    \# Please generate a similar but new question according to the Given Question.
    
    You have four principles to do this.
    
    \# Ensure the new question only asks for one thing, be reasonable, be based on the Given Question, and can be answered with only a number(float or integer). For example, DO NOT ask, ‘what is the amount of A, B and C?’. 
    
    \# Ensure the new question is in line with common sense of life. For example, the amount someone has or pays must be a positive number, and the number of people must be an integer.
    
    \# Ensure your student can answer the new question without the given question. If you want to use some numbers, conditions or background in the given question, please restate them to ensure no information is omitted in your new question.
    
    \# You only need to create the new question. Please DO NOT solve it.
    
    Given Question: {given question}
    
    Your output should be in the following format:
    
    CREATED QUESTION: <your created question>
}
\end{user_example}

\clearpage\onecolumn
\section{Additional Results}
\label{sec:appendix_results}

\begin{figure*}[!h]
    \captionsetup{justification=raggedright, singlelinecheck=false} 
    \caption{
    Xwin-Math's aggregate performance on these two benchmarks is second only to GPT-4, demonstrating our model's robust generalization capabilities.}
    \label{fig:hungarian}
    \centering
    \includegraphics[width=0.7\textwidth]{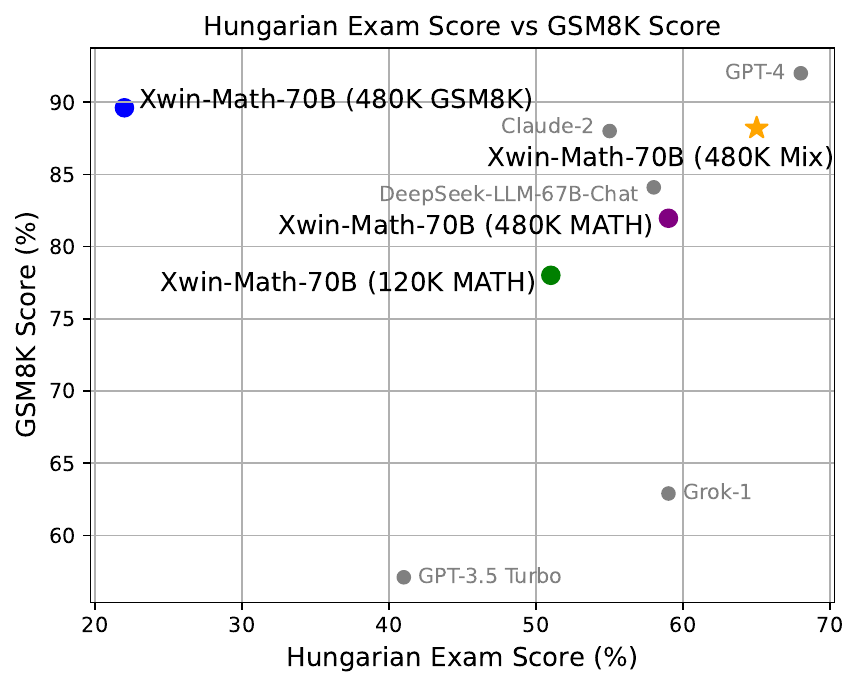}
\end{figure*}

\begin{table*}[!h]
    \caption{
    To validate benchmark data leakage during data generation, we compare LM loss on: 1) a training subset, which is a small subset with 256 samples from all synthetic data; 2) a regenerated training subset, where we maintain the original questions from the training subset and use GPT-4 Turbo to rewrite answers; 3) a regenerated test set, where we keep the questions from the evaluation metrics unchanged and use GPT-4 Turbo to rewrite answers; 4) a reference test set, where we use the test set as seed to generate new questions and answers via GPT-4 Turbo. Referring to Skywork, we also report two key metrics: $\Delta_1=L_{\text{test-regen}}-L_{\text{test-ref}}, \Delta_2=L_{\text{test-regen}}-L_{\text{train-regen}}$,  As $\Delta_1$ is close to 0 and $\Delta_2$ is significantly greater than 0 in two benchmarks, we believe that there is no leakage during the process of data synthesis.
    }
    \label{tab:leakage}
    \centering
    \begin{tabular}{l|c|c|c|c|c|c}
        \toprule
        Dataset & $L_{\text{test-regen}}$ & $L_{\text{test-ref}}$ & $L_{\text{train}}$ & $L_{\text{train-regen}}$ & $\Delta_1$ & $\Delta_2$  \\
        \midrule
        GSM8K & 0.52 & 0.50 & 0.11 & 0.33 & 0.02  & 0.19 \\
        MATH & 0.59 & 0.58 & 0.23 & 0.39 & 0.01 & 0.20 \\ 
        \bottomrule
    \end{tabular}
\end{table*}

\clearpage\onecolumn
\section{Case Study of Synthetic Data}
\label{sec:appendix_verification}
\subsection{GSM8K}

\begin{user_example}[frametitle={Original Question}]
    Randy has 60 mango trees on his farm. He also has 5 less than half as many coconut trees as mango trees. How many trees does Randy have in all on his farm?
\end{user_example}

\begin{user_example}[frametitle={Synthetic Question w/o Verification}]
    Maria has \textcolor{red}{80} apple trees in her orchard. She also has 4 less than a third as many cherry trees as apple trees. How many trees does Maria have in all in her orchard?
\end{user_example}

\begin{user_example}[frametitle={Verification Comment}]
    Let's check the created question:

    - Maria has 80 apple trees.
    
    - She has 4 less than a third as many cherry trees as apple trees. A third of the apple trees is 80/3, which is approximately 26.67. 
    
    - Four less than that would be around 22.67, \textcolor{red}{but since we cannot have a fraction of a tree, we round it to the nearest whole number, which is 27.}
    
    - The modified question should ask for the number of cherry trees without the decimal. So we adjust the question to have a whole number as a result.
    
    - \textcolor{red}{Since 80 is not divisible by 3, to get a whole number of cherry trees, we can change the number so that it is divisible by 3. Let's use 81 apple trees instead of 80.}
\end{user_example}

\begin{user_example}[frametitle={Synthetic Question with Verification}]
    Maria has \textcolor{red}{81} apple trees in her orchard. She also has 4 less than a third as many cherry trees as apple trees. How many trees does Maria have in all in her orchard?
\end{user_example}

\begin{user_example}[frametitle={Synthetic Solution and Final Answer}]
    First, let's determine the number of cherry trees Maria has. We are told she has 4 less than a third as many cherry trees as apple trees.
    
    Since Maria has 81 apple trees, let's find one third of that number:
    
    1/3 of 81 apple trees = 81 apple trees / 3 = 27 apple trees.
    
    Now, since she has 4 less than this amount in cherry trees, we subtract 4 from 27 to find the number of cherry trees:
    
    27 cherry trees - 4 = 23 cherry trees.
    
    To find the total number of trees Maria has in her orchard, we add the number of apple trees to the number of cherry trees:
    
    81 apple trees + 23 cherry trees = 104 trees.
    
    The answer is \fbox{104}.
\end{user_example}

\clearpage
\subsection{MATH}

\begin{user_example}[frametitle={Original Question}]
    The value of $\log_{10}{28471}$ is between the consecutive integers $a$ and $b$.  Find $a+b$.
\end{user_example}

\begin{user_example}[frametitle={Synthetic Question w/o Verification}]
    The value of \textcolor{red}{$\log_{2}{16384}$} is between the consecutive integers $x$ and $y$. Find $x+y$.
\end{user_example}

\begin{user_example}[frametitle={Verification Comment}]
    To ensure the question can be solved, we need to check if the logarithm given can be evaluated and falls between two consecutive integers.
    
    1. Identify the value of $2^n$ that equals $16384$.
    
    2. If $16384$ is a power of $2$, then $\log_{2}{16384}$ will be an integer.
    
    3. Check to ensure the result is an integer and find the consecutive integers around it.
    
    Let's perform the check:
    
    $2^0 = 1$, $2^1 = 2$, $2^2 = 4$, $2^3 = 8$, $2^4 = 16$
    
    ...
    
    $2^{14} = 16384$
    
    So, $\log_{2}{16384} = 14$.
    
    \textcolor{red}{The consecutive integers around 14 are (13 and 14) or (14 and 15).}
    
    \textcolor{red}{Since the logarithm gives an exact integer, to make the question valid, we will modify it to use a number that is not an exact power of 2.}
    
    MODIFIED CREATED QUESTION:
    
    The value of $\log_{2}{10000}$ is between the consecutive integers $m$ and $n$. Find $m+n$.
    
    Now, let's check the modified question:
    
    1. Since 10000 is not a power of 2, $\log_{2}{10000}$ will not be an integer.
    
    2. The goal is to find two consecutive integers between which this logarithm value lies.
    
    3. Knowing that $2^{13} = 8192$ and $2^{14} = 16384$, $\log_{2}{10000}$ is between 13 and 14.
    
    The consecutive integers around $\log_{2}{10000}$ are 13 and 14.
\end{user_example}

\begin{user_example}[frametitle={Synthetic Question with Verification}]
    The value of \textcolor{red}{$\log_{2}{10000}$} is between the consecutive integers $m$ and $n$. Find $m+n$.
\end{user_example}

\begin{user_example}[frametitle={Synthetic Solution and Final Answer}]
    To find the consecutive integers between which $\log_{2}{10000}$ lies, we need to find the powers of $2$ that are immediately less than and greater than $10000$.
    
    First, let's find the largest power of $2$ that is less than $10000$. Starting with $2^1, 2^2, \ldots$ and so on, we find:
    
    $2^{10} = 1024$, $2^{11} = 2048$, $2^{12} = 4096$, $2^{13} = 8192$, $2^{14} = 16384$
    
    Here, we see that $2^{13} < 10000 < 2^{14}$, which means that $\log_{2}{10000}$ is greater than $13$ but less than $14$.
    
    Therefore, the consecutive integers $m$ and $n$ between which $\log_{2}{10000}$ lies are $m = 13$ and $n = 14$.
    
    Now we will add these two integers to find the required sum:
    
    $m + n = 13 + 14 = 27$.
    
    The answer is \fbox{27}.
\end{user_example}

\clearpage\onecolumn
\section{Case Study of Model Response}
\label{sec:appendix_response}
\subsection{GSM8K}

\begin{user_example}[frametitle={Question and Ground Truth Solution}]
{
    \fontsize{11pt}{11pt}\selectfont \textbf{Question: }A treasure hunter found a buried treasure chest filled with gems. There were 175 diamonds, 35 fewer rubies than diamonds, and twice the number of emeralds than the rubies.   How many of the gems were there in the chest?

    \textbf{Solution: }
    There were 175 - 35 = 140 rubies in the treasure chest. There were 140 x 2 = 280 emeralds in the treasure chest. So, the total number of gems in the treasure chest is 175 + 140 + 280 = \fbox{595}.
}
\end{user_example}

\begin{user_example}[frametitle={Response -- Xwin-Math-7B (7.5k data) Greedy Decode}]
{
    \fontsize{11pt}{11pt}\selectfont First, let's find out the number of rubies. We are told there are 35 fewer rubies than diamonds. Since there are 175 diamonds, we can calculate the number of rubies as follows:
    
    Number of rubies = Number of diamonds - 35 = 175 - 35 = 140

    Next, we need to find out the number of emeralds. We are told there are twice the number of emeralds than the rubies. So we can calculate the number of emeralds as follows:
    
    Number of emeralds = 2 × Number of rubies = 2 × 140 = 280
    
    Now, we can find the total number of gems by adding the number of diamonds, rubies, and emeralds together:
    
    Total number of gems = Number of diamonds + Number of rubies + Number of emeralds = \dbox{175 + 140 + 280 = 605} \tikzmark{incorrect_gsm8k} 
    
    The answer is: \fbox{605}. \textcolor{red}{\textbf{Incorrect}}
    \begin{tikzpicture}[overlay, remember picture]  
        \draw[-{Stealth[length=2mm]}, thick] (pic cs:incorrect_gsm8k)++(0,-0.1) -- ++(3,-0.4) node[right] {\textcolor{red}{\textbf{Wrong Calculation}}};  
    \end{tikzpicture}
}
\end{user_example}

\begin{user_example}[frametitle={Response -- Xwin-Math-7B (7.5k data) One Sample from Temperature=0.7}]
{
    \fontsize{11pt}{11pt}\selectfont Let's find the number of each type of gem in the chest step by step.
    
    First, we are given the number of diamonds:
    
    Number of diamonds = 175
    
    Next, we are told there are 35 fewer rubies than diamonds. To find the number of rubies, we subtract 35 from the number of diamonds:
    
    Number of rubies = Number of diamonds - 35 = 175 - 35  = 140
    
    Then, we are told there are twice the number of emeralds than the rubies. To find the number of emeralds, we multiply the number of rubies by 2:
    
    Number of emeralds = 2 * Number of rubies = 2 * 140 = 280
    
    Now, we add the number of diamonds, rubies, and emeralds together to find the total number of gems in the chest:
    
    Total number of gems = Number of diamonds + Number of rubies + Number of emeralds = 175 + 140 + 280 = 595
    
    The answer is: \fbox{595}. \textcolor{red}{\textbf{Correct}}
}
\end{user_example}

\begin{user_example}[frametitle={Response -- Xwin-Math-7B (960k data) Greedy Decode}]
{
    \fontsize{11pt}{11pt}\selectfont Let's break down the information given in the problem step by step.

    1. We know there are 175 diamonds.
    
    2. There are 35 fewer rubies than diamonds. Since there are 175 diamonds, we can calculate the number of rubies by subtracting 35 from 175.
    
    Number of rubies = Number of diamonds - 35 = 175 - 35 = 140

    3. There are twice the number of emeralds than the rubies. We can find the number of emeralds by multiplying the number of rubies by 2.

    Number of emeralds = 2 × Number of rubies = 2 × 140 = 280

    Now, to find the total number of gems in the chest, we add the number of diamonds, rubies, and emeralds together.
    
    Total number of gems = Number of diamonds + Number of rubies + Number of emeralds = 175 + 140 + 280 = 595

    The answer is: \fbox{595}. \textcolor{red}{\textbf{Correct}}
}
\end{user_example}

\clearpage
\subsection{MATH}

\begin{user_example}[frametitle={Question and Ground Truth Solution}]
    \textbf{Question: }If $f(x) = \frac{3x-2}{x-2}$, what is the value of $f(-2) +f(-1)+f(0)$? Express your answer as a common fraction.

    \textbf{Solution: }
    $f(-2)+f(-1)+f(0)=\frac{3(-2)-2}{-2-2}+\frac{3(-1)-2}{-1-2}+\frac{3(0)-2}{0-2}=\frac{-8}{-4}+\frac{-5}{-3}+\frac{-2}{-2}=2+\frac{5}{3}+1=$\fbox{$\frac{14}{3}$}
\end{user_example}

\begin{user_example}[frametitle={Response -- Xwin-Math-7B (7.5k data) Greedy Decode}]
    To find the value of $f(-2) +f(-1)+f(0)$, we need to evaluate the function $f(x)$ at these three values of $x$.
    
    First, let's find $f(-2)$:
    $f(x) = \frac{3x-2}{x-2}$
    Plug in $x = -2$:
    $f(-2) = \frac{3(-2)-2}{-2-2} = \frac{-6-2}{-4} = \frac{-8}{-4}$
    
    Now, let's find $f(-1)$:
    Plug in $x = -1$:
    $f(-1) = \frac{3(-1)-2}{-1-2} = \frac{-3-2}{-3} = \frac{-5}{-3}$
    
    Finally, let's find $f(0)$:
    Plug in $x = 0$:
    $f(0) = \frac{3(0)-2}{0-2} = \frac{-2}{-2} = 1$
    
    Now, we can find the sum:
    $f(-2) +f(-1)+f(0) = \frac{-8}{-4} + \frac{-5}{-3} + 1$
    
    To add these fractions, we need to find a common denominator, which is the least common multiple (LCM) of the \dbox{three denominators (-4, -2, and -3)} \tikzmark{unreasonable}. The LCM of these denominators is -12.
    
    Now, we can rewrite each fraction with the common denominator:
    
    \dbox{$\frac{-8}{-4} = \frac{4}{3}$} \tikzmark{incorrect_math_1}, $\frac{-5}{-3} = \frac{5}{3}$, $1 = \frac{12}{12}$

    Now, we can add the fractions:
    \dbox{$\frac{4}{3} + \frac{5}{3} + \frac{12}{12} = \frac{4+5+12}{3} = \frac{19}{3}$} \tikzmark{incorrect_math_2}
    
    The answer is: \fbox{$\frac{19}{3}$}. \textcolor{red}{\textbf{Incorrect}}
    
    \begin{tikzpicture}[overlay, remember picture]  
        \draw[-{Stealth[length=2mm]}, thick] (pic cs:incorrect_math_1)++(0,-0.1) -- ++(5,-1.6) node[right] (box) {\textcolor{red}{\textbf{Wrong calculation}}};
        \draw[-{Stealth[length=2mm]}, thick] (pic cs:incorrect_math_2)++(-1.2,-0.3) -- (box);
    \end{tikzpicture} 
    \begin{tikzpicture}[overlay, remember picture]  
        \draw[-{Stealth[length=2mm]}, thick] (pic cs:unreasonable)++(0,-0.1) -- ++(3,-1.2) node[right] {\textcolor{red}{\textbf{Unreasonable}}};  
    \end{tikzpicture} 
\end{user_example}

\begin{user_example}[frametitle={Response -- Xwin-Math-7B (7.5k data) One Sample from Temperature=0.7}]
    To find the value of $f(-2) +f(-1)+f(0)$, we first need to evaluate each function at these specific values of x.
    
    For $f(-2)$:
    $f(x) = \frac{3x-2}{x-2}$
    Substitute $x = -2$ into the equation:
    $f(-2) = \frac{3(-2)-2}{-2-2} = \frac{-6-2}{-4}= \frac{-8}{-4} = 2$
    
    For $f(-1)$:
    $f(x) = \frac{3x-2}{x-2}$
    Substitute $x = -1$ into the equation:
    $f(-1) = \frac{3(-1)-2}{-1-2} = \frac{-3-2}{-3} = \frac{-5}{-3} = \frac{5}{3}$
    
    For $f(0)$:
    $f(x) = \frac{3x-2}{x-2}$
    Substitute $x = 0$ into the equation:
    $f(0) = \frac{3(0)-2}{0-2} = \frac{-2}{-2} = 1$
    
    Now we can add the values of $f(-2)$, $f(-1)$, and $f(0)$:
    $f(-2) +f(-1)+f(0) = 2 + \frac{5}{3} + 1 = \frac{14}{3}$
    
    The answer is: \fbox{$\frac{14}{3}$}. \textcolor{red}{\textbf{Correct}}
\end{user_example}

\begin{user_example}[frametitle={Response -- Xwin-Math-7B (480k data) Greedy Decode}]
    To find the value of $f(-2) + f(-1) + f(0)$, we need to evaluate the function $f(x)$ at each of the given points and then sum the results.
    
    First, let's find $f(-2)$:
    $f(-2) = \frac{3(-2) - 2}{(-2) - 2} = \frac{-6 - 2}{-2 - 2} = \frac{-8}{-4} = 2$
    
    Next, let's find $f(-1)$:
    $f(-1) = \frac{3(-1) - 2}{(-1) - 2} = \frac{-3 - 2}{-1 - 2} = \frac{-5}{-3} = \frac{5}{3}$
    
    Finally, let's find $f(0)$:
    $f(0) = \frac{3(0) - 2}{0 - 2} = \frac{0 - 2}{-2} = \frac{-2}{-2} = 1$
    
    Now, we can sum these values:
    $f(-2) + f(-1) + f(0) = 2 + \frac{5}{3} + 1$
    To add these, we need a common denominator, which is 3:
    $2 + \frac{5}{3} + 1 = \frac{2 \cdot 3}{3} + \frac{5}{3} + \frac{1 \cdot 3}{3} = \frac{6}{3} + \frac{5}{3} + \frac{3}{3}$
    
    Now, we add the numerators:
    $\frac{6}{3} + \frac{5}{3} + \frac{3}{3} = \frac{6 + 5 + 3}{3} = \frac{14}{3}$
    
    The answer is: \fbox{$\frac{14}{3}$}. \textcolor{red}{\textbf{Correct}}
\end{user_example}

\end{document}